\documentclass[letterpaper]{article} 
\usepackage[preprint]{aaai2027}  
\usepackage[hyphens]{url}  
\usepackage{graphicx} 
\usepackage{natbib}  
\usepackage{caption} 
\usepackage{algorithm}
\usepackage{algorithmic}
\usepackage{amsmath}
\usepackage{amssymb}
\usepackage{amsfonts}
\usepackage{bm}
\usepackage{booktabs}
\usepackage{comment}
\usepackage{makecell}  
\definecolor{pearDark}{HTML}{2980B9}
\usepackage{capt-of}

\newcommand{\vt}{v_{\theta}}
\newcommand{\vold}{v^{\mathrm{old}}}
\newcommand{\told}{\theta^{\mathrm{old}}}
\newcommand{\vcfg}{v^{\mathrm{cfg}}_{\mathrm{old}}}
\newcommand{\uvec}{u_t}
\newcommand{\Adv}{\mathcal{A}}
\newcommand{\Advs}{\mathcal{A}_{s}}
\newcommand{\vs}{v_{s}}
\newcommand{\Advhat}{\widehat{\mathcal{A}}}
\newcommand{\pold}{p^{\mathrm{old}}}
\newcommand{\qs}{q_{s}}

\newcommand{\Exp}{\mathbb{E}}
\newcommand{\Rew}{R}

\newtheorem{proposition}{Proposition}
\newcommand{\vsstar}{v_{s}^{*}}

\newcommand{\teasercaption}{%
\textbf{Why cold-start video RL needs guided rollouts.}
\textit{Left:} rollouts from Wan2.1-1.3B on two prompts, with and without CFG,
annotated with HPSv3 rewards $R^i$. Without CFG every rollout is corrupted
(smearing, broken structure) yet the rewards remain widely spaced ($-3.42$ to
$-10.12$ on the first prompt): these intra-group gaps track no perceptible
quality difference and cannot serve as a relative signal.
\textbf{(a)} Unguided samples concentrate where the reward model is least
discriminative (per-prompt median $-0.47$ vs.\ $4.72$).
\textbf{(b)} Without CFG, the reward first drifts \emph{downward} and recovers only after ${\sim}150$ GPU hours, while
guided rollouts grow stably.%
}
\title{VideoAWR: Reinforcing Video Flow Models with\\ Decoupled-Timestep Advantage-Weighted Rectification}

\author{
    Zheming Fu\textsuperscript{1,2,3}\equalcontrib,
    Wenbin Ning\textsuperscript{1}\equalcontrib,
    Ruizhe He\textsuperscript{3,4},
    Wei Shang\textsuperscript{1},
    Xiaoxiao Ma\textsuperscript{3,5},\\
    Lianrui Mu\textsuperscript{3,4},
    Chang Liu\textsuperscript{2}\corresponding,
    Lei Wang\textsuperscript{1}\corresponding,
    Siming Fu\textsuperscript{3}\corresponding
}
\affiliations{
    \textsuperscript{1}Beihang University,
    \textsuperscript{2}Zhongguancun Academy,
    \textsuperscript{3}Joy Future Academy\\
    \textsuperscript{4}Zhejiang University,
    \textsuperscript{5}University of Science and Technology of China
}

\makeatletter
\def\@maketitle{%
  \def\theauthors{\if T\showauthors@on\@author\else Anonymous submission\fi}%
  \newcounter{eqfn}\setcounter{eqfn}{0}%
  \newcounter{corrfn}\setcounter{corrfn}{0}%
  \vbox{%
    \let\footnote\thanks\relax%
    \setcounter{footnote}{0}%
    \def\equalcontrib{%
      \ifnum\value{eqfn}=0%
        \footnote{These authors contributed equally.}%
        \setcounter{eqfn}{\value{footnote}}%
      \else%
        \footnotemark[\value{eqfn}]%
      \fi%
    }%
    \def\corresponding{%
      \ifnum\value{corrfn}=0%
        \footnote{Corresponding authors.}%
        \setcounter{corrfn}{\value{footnote}}%
      \else%
        \footnotemark[\value{corrfn}]%
      \fi%
    }%
    \hsize\textwidth%
    \linewidth\hsize%
    \vskip 0.625in minus 0.125in%
    \centering%
    {\LARGE\bf \@title \par}%
    \vskip 0.1in plus 0.5fil minus 0.05in%
    {\Large{\textbf{\theauthors\ifhmode\\\fi}}}%
    \vskip .2em plus 0.25fil%
    {\normalsize \aaai@affiliations\ifhmode\\\fi}%
    \vskip 0.7em%
    \includegraphics[width=0.90\textwidth]{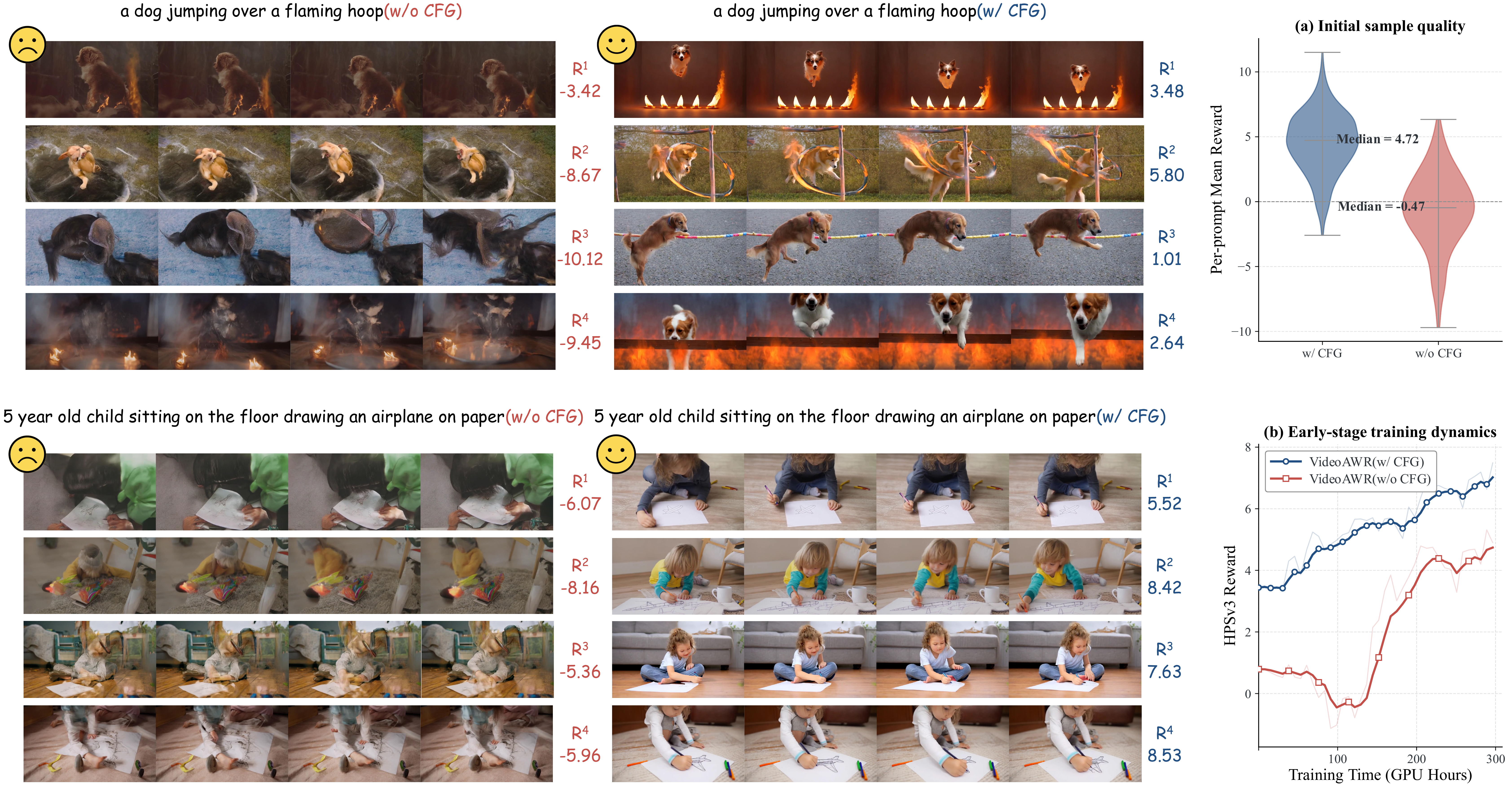}\par%
    \refstepcounter{figure}\label{fig:motivation}%
    \noindent\makebox[\textwidth][c]{%
      \begin{minipage}{0.96\textwidth}%
      \small {Figure~\thefigure:} \teasercaption%
      \end{minipage}%
    }\par%
    \vskip 1em plus 2fil%
  }%
}%
\makeatother

\begin{document}

\maketitle

\begin{abstract}
Regression-based RL for flow models mandates sampling \emph{without} classifier-free guidance (CFG) to align the rectification target with the reference marginal velocity. Extending Advantage-Weighted Rectification (AWR), which dynamically scales this regression target based on reward, to video generation encounters a cold-start pathology: unguided rollouts collapse into low-fidelity regions where intra-group reward gaps remain wide but no longer track perceptual quality, rendering the optimization target arbitrary. Conversely, utilizing guided rollouts to restore fidelity, while keeping the regression target tied to the guided sampling distribution, provably triggers a geometric guidance blow-up under exponential-moving-average (EMA) updates. We propose \textbf{VideoAWR}, which resolves this tension by untying sampling from the optimization anchor. An EMA reference policy generates \emph{guided} rollouts to restore reward separability, while the trainable policy is rectified against the \emph{unguided} conditional anchor. We prove this decoupled objective is mathematically exact: the centered advantage cancels the anchor in expectation, filtering out the reward-neutral guidance offset and retaining only the reward-correlated covariance increment. Consequently, inference runs single-branch and CFG-free, yet inference-time CFG preserves its original semantics. Furthermore, we introduce \emph{Decoupled Timestep Rectification}, supervising each rollout at an independently sampled timestep without bias to improve sample efficiency. Experiments on Wan2.1-1.3B and 14B demonstrate stable optimization and superior visual and motion quality in \emph{both} CFG-free and CFG-enabled regimes.
Code is available at \pdfstartlink attr{/Border [0 0 0]}
user{/Subtype /Link /A << /S /URI /URI (https://github.com/Fuzm7/FlowAWR) >>}%
\textcolor{pearDark}{VideoAWR}\pdfendlink.
\end{abstract}

\section{Introduction}
Reinforcement learning for generative alignment~\citep{black2024ddpo, fan2023dpok, rafailov2023direct} extends to text-to-video~\citep{prabhudesai2024video, liu2024videodpo, xing2024survey} primarily through policy-gradient~\citep{shao2024deepseekmath, liu2026flow, xue2025dancegrpo} and regression-based methods~\citep{zheng2025diffusionnft, fu2026flowawr}. The latter embeds the reward directly into the velocity-matching objective~\citep{peters2010relative, peng2019awr}, bypassing intractable trajectory-likelihood estimation over long rollouts. However, regression dictates strict data collection constraints: the target anchor---the marginal velocity of a reference path~\citep{lipman2022flow, liu2023flow}---doubles as the sampling field. Because CFG~\citep{ho2022cfg, dhariwal2021diffusion} distorts this marginal, rollouts must theoretically remain \emph{unguided}.

In video generation, this unguided requirement degrades optimization. Sampled without CFG, models like Wan2.1~\citep{wan2025} yield videos confined to a low-fidelity region where reward models~\citep{wu2023hps, ma2025hpsv3, liu2026improving} lose discriminative capacity (Figure~\ref{fig:motivation}). The induced intra-group ranking becomes arbitrary, causing initial reward degradation. Conversely, restoring fidelity via guided rollouts~\citep{guo2025g, frans2025diffusion} while anchoring on that same CFG field induces inter-round instability. We prove that under EMA updates, this coupling geometrically compounds the conditional--unconditional gap. Must guided exploration then be paid for with unstable training?

We propose \textbf{VideoAWR}, which explicitly separates the \emph{sampling distribution} from the \emph{optimization anchor}. The EMA reference policy generates \emph{guided} rollouts to maintain fidelity and reward separability, while the trainable policy is rectified against the \emph{unguided} conditional anchor. Characterizing the population optimum of this decoupled objective demonstrates that the mismatch is mathematically exact: the centered advantage cancels the anchor in expectation. This isolates the unguided conditional velocity and the reward-correlated covariance increment, \textbf{strictly filtering out the guidance offset}. Consequently, inference operates via a single forward branch, and the conditional--unconditional gap shifts only by an advantage-bounded increment, preserving inference-time CFG semantics. This structurally contrasts with guidance distillation~\citep{meng2023distillation, tang2025diffusion}, which forfeits inference-time guidance by baking the field directly into the network.

Furthermore, the point-wise evaluation of the regression objective decouples the training timestep from the rollout schedule, bypassing limitations of single-step schemes that tie timesteps to the sampling discretization or share them across groups~\citep{li2025mixgrpo, he2026flash, tempflowgrpo2025}. We introduce \textbf{Decoupled Timestep Rectification}, assigning an independent timestep per rollout without bias. The exact importance correction collapses to a scalar naturally absorbed by group normalization, a structural flexibility unique to regression objectives. Our contributions are as follows:
\begin{itemize}
    \item \textbf{Guided rollout with a stable anchor.} We identify the cold-start pathology of unguided video flow RL and prove that its conventional remedy---anchoring the target on the guided CFG field---diverges. Under EMA updates, the gap between the conditional and unconditional fields compounds geometrically.

    \item \textbf{Anchor invariance of the behavior-sampled optimum.} We resolve this divergence by separating the guided sampling distribution from the unguided conditional anchor. We prove its population optimum is exactly the unguided field plus the reward-driven increment; the centered advantage strictly filters out the guidance offset in expectation. This inherently enables CFG-free inference while preserving inference-time guidance semantics.

    \item \textbf{Decoupled Timestep Rectification.} We introduce an unbiased per-rollout timestep estimator that improves sample efficiency by allowing a single prompt group to supervise multiple timesteps. We show its residual importance correction collapses to a scalar natively absorbed by group normalization.

    \item \textbf{Empirical validation under matched compute.} Evaluated under equal GPU-hours on Wan2.1-1.3B and 14B, VideoAWR achieves higher visual and motion rewards than single-step baselines, optimizes stably throughout, and retains its gains under both CFG-free and CFG-enabled inference.
\end{itemize}

\section{Related Work}
\label{sec:related}

\paragraph{Alignment for Video Generation.}
Following the success of RLHF in language and image domains~\citep{ouyang2022training, black2024ddpo, fan2023dpok}, aligning text-to-video models with human preferences has drawn increasing attention. Direct preference optimization has been extended to video by constructing preference pairs at different granularities~\citep{liu2024videodpo, wu2026densedpo, zhang2024onlinevpo}. Parallelly, online RL formulations treat video diffusion and flow processes as multi-step Markov decision processes, optimizing reverse-process transitions via policy gradients~\citep{black2024ddpo, fan2023dpok}. For continuous flow-based video models, Flow-GRPO~\citep{liu2026flow} and DanceGRPO~\citep{xue2025dancegrpo} introduce stochasticity into ODE integration to yield tractable transition probabilities and enable group-relative credit assignment. To mitigate the heavy computational overhead of long spatiotemporal trajectories, recent extensions adopt mixed ODE--SDE sampling or single-timestep trajectory supervision~\citep{li2025mixgrpo, he2026flash}. However, these policy-gradient derivatives struggle when the unguided generation quality is severely degraded at cold start.

\paragraph{Classifier-Free Guidance and Flow Fine-tuning.}
Classifier-free guidance boosts conditional alignment by extrapolating conditional velocity fields away from unconditional ones, at the expense of an additional evaluation branch~\citep{ho2022cfg}; guidance distillation avoids this overhead by regressing the combined field into a single network~\citep{meng2023distillation}. For flow-based preference alignment, DiffusionNFT~\citep{zheng2025diffusionnft} models positive and negative transitions on the forward process, while FlowAWR~\citep{fu2026flowawr} anchors advantage-weighted rectification on the reference policy's velocity field. These methods work for image generation without CFG at sampling time, but in video they meet the cold-start pathology above. VideoAWR instead confines CFG to the sampler.

\section{Preliminaries}

\begin{figure*}[!t]
    \centering
    \includegraphics[width=0.98\textwidth]{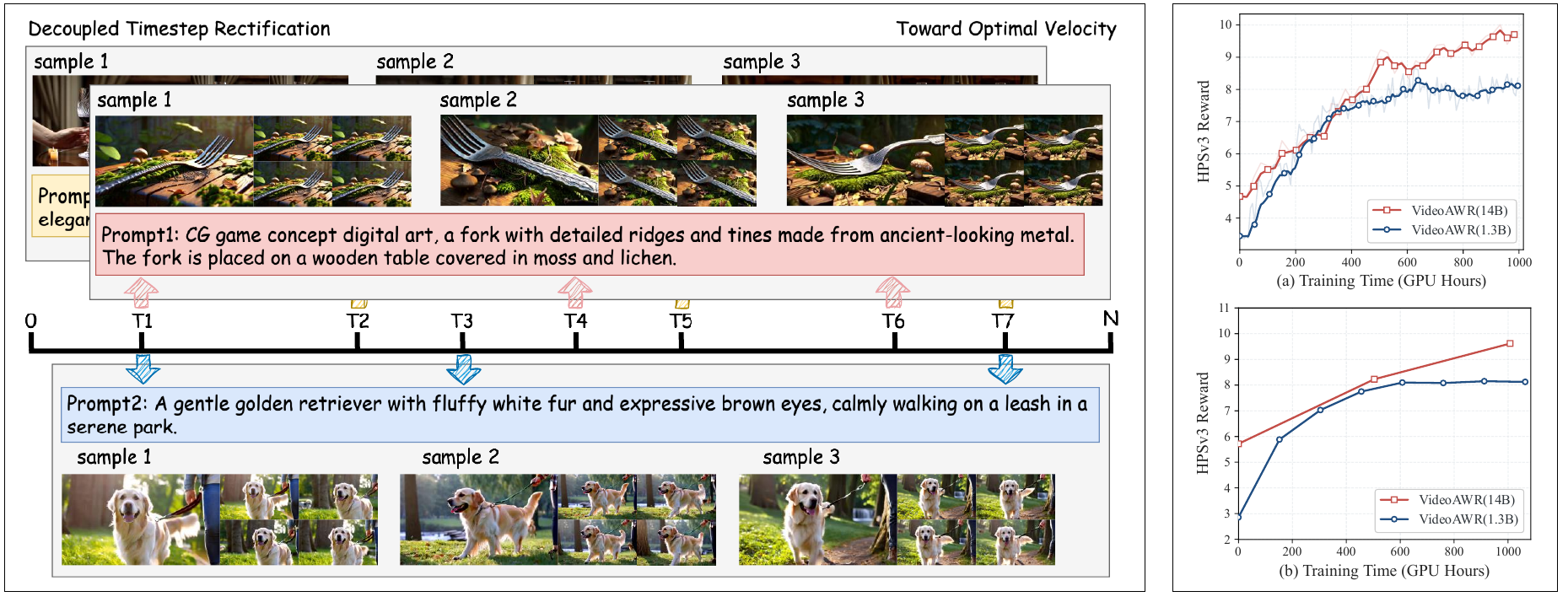}
    \vspace{-0.5em}
    \caption{\textbf{Decoupled Timestep Rectification and scaling to 14B.}
    \textit{Left:} DTR assigns each rollout in a prompt group its own
    independently drawn timestep on $[0,N]$, so one group supervises the
    velocity field at several distinct points instead of a single shared
    timestep (Sec.~\ref{sec:dtr}). \textit{Right:} HPSv3 reward of VideoAWR on
    Wan2.1-1.3B and Wan2.1-14B against training cost in GPU hours, for
    \textbf{(a)} training and \textbf{(b)} the held-out evaluation prompts.
    Both backbones improve under the same hyperparameters.}
    \label{fig:pipeline}
    \vspace{-1em}
\end{figure*}

\subsection{Flow Matching and Rectified Flow}
\label{subsec:prelim_fm}
Given a noise prior $p_0=\mathcal{N}(0,I)$ and a conditional data distribution $p_1(\cdot\mid c)$ over video latents with prompt $c$, Flow Matching~\citep{lipman2022flow} constructs a probability path transporting $p_0$ to $p_1$. Under the Rectified Flow formulation~\citep{liu2023flow}, the forward process interpolates linearly:
\begin{equation*}
    x_t=(1-t)\,x_0+t\,x_1,
\end{equation*}
which induces the constant conditional velocity target $\uvec(x_t\mid x_1)=x_1-x_0=\frac{x_1-x_t}{1-t}$. A parameterized network $\vt(x_t,c,t)$ is trained to regress this target:
\begin{equation}
    \mathcal{L}_{\mathrm{FM}}(\theta)=\Exp_{t,\,x_0,\,x_1}\big[\|\vt(x_t,c,t)-\uvec(x_t\mid x_1)\|^2\big],
    \label{eq:fm_loss}
\end{equation}
with $t\sim\mathcal{U}[0,1]$, $x_0\sim p_0$, $x_1\sim p_1(\cdot\mid c)$. Its global minimum is the marginal velocity field $v(x_t,c,t)=\Exp_{p(x_1\mid x_t,c)}[\uvec(x_t\mid x_1)]$, i.e.\ the posterior mean of the conditional targets~\citep{lipman2022flow}. Samples are generated by integrating the deterministic ODE $\mathrm{d} x_t=\vt(x_t,c,t)\,\mathrm{d} t$ from $x_0\sim p_0$~\citep{chen2018neural}.

\subsection{Reinforcement Learning as Flow Fine-tuning}
\label{subsec:prelim_ft}
Since flow-based generative models sample trajectories via a deterministic ODE, direct computation of trajectory likelihoods is intractable, which necessitates SDE reformulations of the generative process for policy gradient methods~\citep{liu2026flow, xue2025dancegrpo}. Alternatively, a regression-based paradigm avoids trajectory-level likelihood estimation by embedding the reinforcement signal directly into the velocity matching objective. For instance, DiffusionNFT~\citep{zheng2025diffusionnft} constructs implicit positive and negative policies by mixing the trainable field $\vt$ with the reference field $\vold$ via a scaling $\beta$, which is formulated as a reward-weighted regression problem:
\begin{equation}
    \label{eq:nft_loss}
    \begin{aligned}
        \mathcal{L}_{\mathrm{NFT}}(\theta) &\propto\Exp\Big[\,
        r\,\big\|\vt- \left(\vold + \tfrac{1}{\beta}(\uvec-\vold)\right)\big\|^2 \\
        &+\,(1-r)\,\big\|\vt- \left(\vold-\tfrac{1}{\beta}(\uvec-\vold) \right) \big\|^2\Big],
    \end{aligned}
\end{equation}
where $r\in[0,1]$ denotes the normalized terminal scalar reward, and spatiotemporal arguments are omitted for notation brevity. Pointwise minimization of Eq.~\eqref{eq:nft_loss} with respect to $\vt$ yields the optimal target $\vold+\frac{2r-1}{\beta}(\uvec-\vold)$. This analytical solution demonstrates that the reference field is modulated along the residual direction $(\uvec-\vold)$ by a reward-dependent weight restricted to the interval $[-1/\beta, 1/\beta]$.

FlowAWR~\citep{fu2026flowawr} extends this discrete formulation into a generalized probabilistic framework. Under a KL-constrained reward maximization formulation with a temperature parameter $\gamma>0$, the optimal terminal density corresponds to the exponentially re-weighted reference distribution $p_1^*(x_1\mid c)\propto p_1^{\mathrm{old}}(x_1\mid c)\exp(\Rew/\gamma)$. The marginal velocity field that induces $p_1^*$ can be explicitly derived as an advantage-weighted rectification of the reference field:
\begin{equation}
    v^*=\vold+\Exp_{\pold(x_1\mid x_t,c)}\big[\Adv\,(\uvec-\vold)\big],
    \label{eq:awr_opt}
\end{equation}
where $\Adv=\exp(\Rew/\gamma)/\phi-1$ represents the centered advantage, and $\phi=\Exp_{\pold(x_1\mid x_t,c)}[\exp(\Rew/\gamma)]$ is the time-dependent value function. In practice, following GRPO, $\phi$ is empirically approximated by the arithmetic mean of a group of $G$ rollouts generated from distinct initial noises $\{x_0^{(i)}\}_{i=1}^G$. For the $i$-th sample within the group, this formulation yields a Softmax-style empirical advantage:
\begin{equation}
    \label{eq:softmax_advantage}
    \Adv^{(i)} \approx \frac{\exp(\Rew(c, x_1^{(i)})/\gamma)}{\frac{1}{G}\sum_{j=1}^G \exp(\Rew(c, x_1^{(j)})/\gamma)} - 1.
\end{equation}
Regressing $\vt$ onto this group-estimated rectified target defines the practical training objective:
\begin{equation}
    \begin{aligned}
    \mathcal{L}_{\mathrm{AWR}}(\theta) = &\Exp_{t, x_0^{(i)}, x_1^{(i)}\sim \pold} \big\| \vt^{(i)} \\
    &- \big( \vold + \Adv^{(i)} ( \uvec^{(i)} - \vold ) \big) \big\|^2.
\end{aligned}
\label{eq:awr_loss}
\end{equation}
Thus, DiffusionNFT can be interpreted as a reward-bounded special case where the adjustment magnitude is constrained within the fixed interval $\pm 1/\beta$, whereas FlowAWR dynamically scales the rectification via the group-relative estimation.

Both targets share the general form $v^{\mathrm{anc}}+\Adv\,(\uvec-v^{\mathrm{anc}})$. We define $v^{\mathrm{anc}}$ as the \emph{anchor} of the rectification: it is detached from $\theta$, defines the fallback trajectory when the advantage vanishes, and, because $\Adv$ is centered, serves as the baseline field retained by the learned policy. In FlowAWR, the anchor $v^{\mathrm{old}}$ simultaneously generates the sampling trajectories.

\section{Method}

VideoAWR adapts the advantage-weighted rectification paradigm to the alignment of text-to-video flow models, which presents two primary challenges.
First, the reference field is structurally mismatched to the reward signal:
dropping CFG destroys the reward separability that
relative advantage estimation depends on, yet anchoring the target on the amplified CFG
field makes training diverge (Sec.~\ref{sec:behavior_alignment}). Second, the
training timestep must be sampled cheaply and without bias, ideally decoupling the temporal supervision from the terminal reward, allowing optimization via point-wise regression to maximize sample efficiency
(Sec.~\ref{sec:dtr}).

\subsection{Behavior-Anchored Flow Alignment}
\label{sec:behavior_alignment}
Applied directly to video generation, FlowAWR exhibits early-stage optimization instability. 
Without CFG, the pretrained backbone produces low-fidelity samples that the reward model cannot reliably order. And the resulting advantages exhibit high variance, yielding marginal initial gains rather than effective alignment. We restore separability by rolling out with a guidance scale $s>1$, which extrapolates the conditional branch away from the unconditional one:
\begin{equation}
\label{eq:cfg-field}
\vcfg(x_t,c)=\vold(x_t,\varnothing)+s\big(\vold(x_t,c)-\vold(x_t,\varnothing)\big).
\end{equation}
Integrating the deterministic ODE induced by $\vcfg$ transports the noise
prior to a guided terminal distribution $\qs(\cdot\mid c)$, whose samples are
separated in reward that tracks perceptible quality, supporting
meaningful relative judgments.

\paragraph{Why not anchoring on the CFG field?}
Guided rollouts raise a design question: which field should serve as the
anchor of the rectification target in the sense of Sec.~\ref{subsec:prelim_ft}? The apparently natural answer is $\vcfg$ itself, so
that the target and the sampler share a single field. This is consistent
within one round but unstable across online updates. Let
$\Delta_k=\vold_{k}(x_t,c)-\vold_{k}(x_t,\varnothing)$ denote the
gap between the conditional and unconditional velocity field at iteration $k$.

\begin{proposition}[Guidance Explosion]
\label{prop:geometric_blowup}
Suppose each step drives the conditional branch to its CFG
anchor while the unconditional branch remains invariant, and suppose the EMA
update $\told_{k+1}\leftarrow\mu_k\told_k+(1-\mu_k)\theta_{k+1}$, which is linear
in parameters, induces a linear update in the field space up to the first order. Then the gap
evolves as
\begin{equation}
\label{eq:blowup}
\begin{aligned}
    \Delta_{k+1} &\approx \big[\mu_k + (1-\mu_k)s\big]\Delta_k \\
    \Rightarrow \Delta_k &\sim \left( \prod_{i=0}^{k-1} \big[\mu_i + (1-\mu_i)s\big] \right) \Delta_0,
\end{aligned}
\end{equation}
where the step-wise factor is determined by the EMA schedule $\mu_k$ and guidance scale $s$. Because the EMA update is linear in parameter space, its first-order Taylor expansion linearly projects this update into the velocity field space, directly preserving the state-independent hyperparameters as the amplification coefficients.
\end{proposition}

Since $\mu_i+(1-\mu_i)s>1$ for any $\mu_i \in (0,1)$ whenever $s>1$, the product strictly diverges, amplifying the gap geometrically as long as the first-order linearization holds. This theoretical divergence corresponds to the empirical observation in Figure~\ref{fig:ablation}(a), where the evaluation reward monotonically degrades from the onset of training, preceding the numerical crash at $300$ GPU hours. The sampler operates at an increasing effective guidance strength, driving the generated video latents $x_t$ off the pre-trained data manifold and inducing training collapse (proof in Appendix~\ref{app:proof_prop1}).

VideoAWR therefore separates \emph{where guidance acts} from \emph{what the
policy learns}: guidance is confined to the behavior sampler, while the
rectification target is anchored on the unguided conditional field. Given a
group of guided rollouts $\{x_1^{(i)}\}_{i=1}^G\sim\qs(\cdot\mid c)$ with
empirical advantages $\Advhat^{(i)}$ from Eq.~\eqref{eq:softmax_advantage},
the behavior-anchored target is
\begin{equation}
\label{eq:target}
y_t^{(i)}=\vold(x_t^{(i)},c)
+\Advhat^{(i)}\big(\uvec(x_t^{(i)}\mid x_1^{(i)})-\vold(x_t^{(i)},c)\big),
\end{equation}
where quantity in $y_t^{(i)}$ is detached from $\theta$. The trainable
policy minimizes the squared error under the guided behavior distribution:
\begin{equation}
\label{eq:loss}
\mathcal{L}_{\mathrm{VideoAWR}}(\theta)=
\Exp_{\substack{t, \ x_0^{(i)}\sim\, p_0\\[1pt]
x_1^{(i)}\sim\,\qs(\cdot\mid c)}}
\Big[\big\|\vt(x_t^{(i)},c,t)-y_t^{(i)}\big\|_2^2\Big].
\end{equation}
Because the anchor $\vold$ and the sampling distribution $\qs$ are deliberately
mismatched, it is not obvious what field this hybrid objective converges
to, a question we settle next.

\begin{figure*}[!t]
    \centering
    \captionof{table}{Detailed comparison of \textbf{General Video Quality} using VBench metrics. We evaluate aesthetic quality, image quality, subject consistency, and object class. We reproduce the official VBench results. Best scores are in \textbf{Bold}.}
    \label{vbench}
    
    \small
    \setlength{\tabcolsep}{12pt}
    \begin{tabular}{l|c|cccc}
    \toprule
    \textbf{Method} & \textbf{GPU Hours} & \makecell[c]{\textbf{Aesthetic}\\\textbf{Quality} $\uparrow$} & \makecell[c]{\textbf{Imaging}\\\textbf{Quality} $\uparrow$} & \makecell[c]{\textbf{Subject}\\\textbf{Consistency} $\uparrow$} & \makecell[c]{\textbf{Object}\\\textbf{Class} $\uparrow$} \\
    \midrule
    CogVideoX-2B~\cite{yang2024cogvideox} & -- & 61.07 & 62.37 & 96.52 & 86.48 \\
    Hunyuan-Video~\cite{kong2024hunyuanvideo} & -- & 60.36 & 67.56 & 97.37 & 86.10 \\
    \midrule
    Wan2.1-T2V-1.3B~\cite{wan2025} & -- & 65.63 & 66.79 & 94.58 & 88.70 \\
    \textbf{Flow-GRPO} & 1100 & 66.05 & 68.34 & 94.99 & 89.59 \\
    \textbf{Flow-GRPO-Fast1} & 600 & 65.66 & 64.19 & 95.45 & 89.85 \\
    \textbf{Flash-GRPO} & 600 & 67.46 & 68.10 & 95.85 & 90.36 \\
    \textbf{DiffusionNFT} & 600 & 68.30 & \textbf{71.50} & 97.49 & 90.90 \\
    \textbf{VideoAWR} & 600 & \textbf{69.18} & 71.01 & \textbf{98.48} & \textbf{92.53} \\
    \bottomrule
    \end{tabular}
    \vspace{2pt} 
    \includegraphics[width=0.98\textwidth]{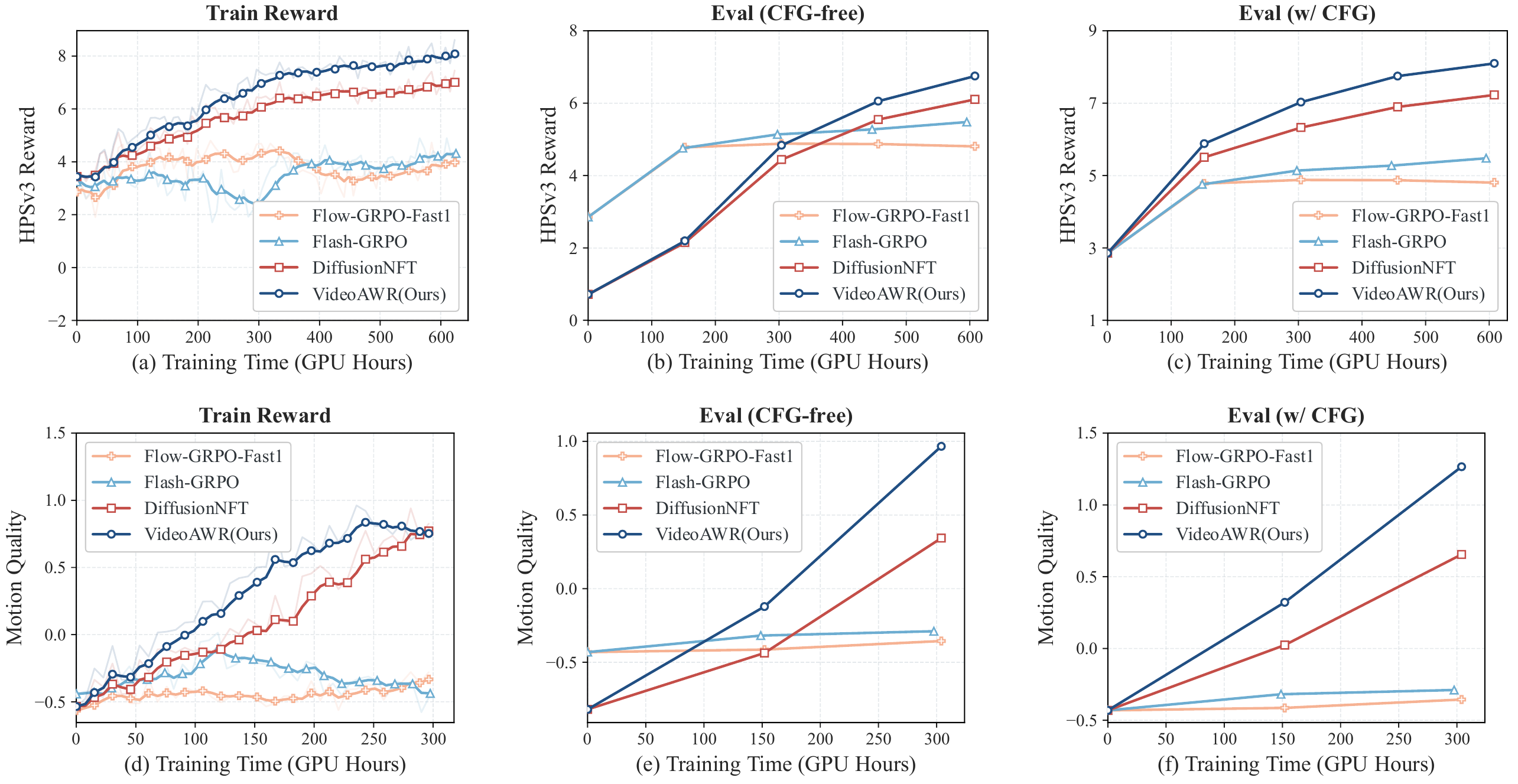}
    \vspace{-1em}
    \captionof{figure}{\textbf{Reward optimization on HPSv3 (top) and Motion Quality
    (bottom) against single-step baselines,} plotted against training cost in
    GPU hours (shaded bands: raw traces). Panels \textbf{(a,\,d)} report
    training reward; evaluation on the held-out $300$ prompts is split into
    \textbf{(b,\,e)~CFG-free} and \textbf{(c,\,f)~with CFG}. VideoAWR attains
    the highest reward on both metrics, while Flow-GRPO-Fast1 and Flash-GRPO
    plateau earlier and DiffusionNFT remains below. The CFG-free panels
    (b,\,e) indicate that the guided-rollout gains are retained by the
    unguided policy; enabling CFG (c,\,f) improves it further.}
    \label{fig:main_curves}
    \vspace{-1em}
\end{figure*}

\subsection{What the Behavior-Anchored Objective Learns}
\label{sec:analysis}
To characterize the population optimum of Eq.~\eqref{eq:loss}, we define the
value function
$\phi_s(x_t,c,t)=\mathbb{E}_{q_{s}(x_1\mid x_t,c)}[\exp(R/\gamma)]$ and the
centered advantage $\mathcal{A}_{s}=\exp(R/\gamma)/\phi_s-1$ under the guided
behavior flow, of which the group $\Advhat^{(i)}$ is the empirical
estimate. Let $v_{s}=\mathbb{E}_{q_{s}(x_1\mid x_t,c)}[u_t]$ be the marginal
behavior velocity, and
$v_{s}^*=v_{s}+\mathbb{E}_{q_{s}}[\mathcal{A}_{s}(u_t-v_{s})]$ the
corresponding optimal field from Eq.~\eqref{eq:awr_opt}.
\begin{proposition}[Anchor-Invariant Improvement]
\label{prop:population_optimum}
For a fixed state $(x_t,c,t)$, the population minimizer $v^{\dagger}$ of
Eq.~\eqref{eq:loss}, with $\Advhat^{(i)}$ replaced by its population limit
$\mathcal{A}_s$, satisfies:
\begin{equation}
\label{eq:gap}
v^{\dagger} = v^{\mathrm{old}} + (v_{s}^* - v_{s}).
\end{equation}
\end{proposition}
The proof (Appendix~\ref{app:proof_prop2}) uses the zero-mean property
$\mathbb{E}_{q_{s}}[\mathcal{A}_{s}]=0$, which makes any $x_1$-independent
anchor immaterial in expectation.

\paragraph{What $v^{\dagger}$ represents.}
The field generating the tilted guided density
$q_{s}^{*}(x_1\mid c)\propto q_{s}(x_1\mid c)\exp(\Rew/\gamma)$ is $v_{s}^*$,
\emph{not} $v^{\dagger}$: the two differ by
$v^{\dagger}-v_{s}^*=v^{\mathrm{old}}-v_{s}$, exactly the reward-neutral
guidance shift whose accumulation drives
Proposition~\ref{prop:geometric_blowup}. VideoAWR filters it out by
construction, leaving
\begin{equation*}
v^{\dagger}=\underbrace{v^{\mathrm{old}}}_{\text{unguided backbone}}
+\underbrace{\mathrm{Cov}_{q_{s}}(\Advs,\uvec\mid x_t,c)}
_{\text{reward-correlated increment}},
\end{equation*}
where the increment follows from $\Exp_{q_{s}}[\Advs]=0$.

This formulation yields two structural properties. First, by preserving the unguided conditional field as the backbone, it enables single-forward-pass inference. Furthermore, the conditional--unconditional gap of the trained policy differs from the base model solely by the advantage-bounded increment, circumventing the compounding effect of $s$. Consequently, inference-time CFG retains its original semantics, allowing the model to serve both regimes.

According to Eq.~\eqref{eq:awr_opt} instantiated under $q_{s}$, the covariance term is precisely the displacement required to shift the behavior flow toward the field of its KL-regularized reward optimum $q_{s}^{*}$. Applying this increment to $v^{\mathrm{old}}$ instead of $v_{s}$ internalizes the reward-driven improvement without importing the guidance offset, allowing the increments to compound robustly under EMA. Since $q_s$ is intrinsically generated by applying CFG to $v^{\mathrm{old}}$, the optimization loop is self-consistent, reaching a theoretical fixed point exactly when $\mathrm{Cov}_{q_{s}}(\Advs,\uvec\mid x_t,c)=0$ (i.e., when guided rollouts can no longer separate high- from low-reward terminals).

\begin{figure*}[!t]
    \centering
    \includegraphics[width=0.99\textwidth]{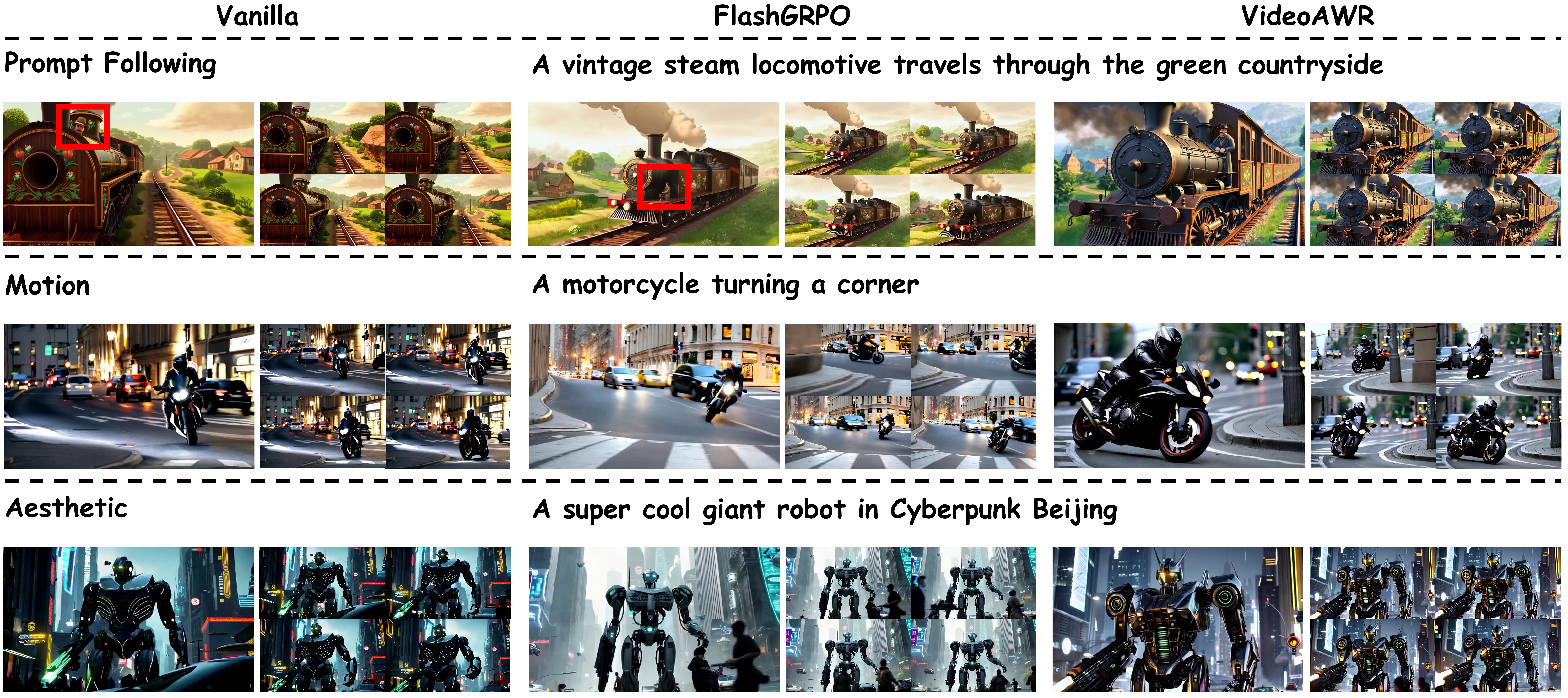}
    \vspace{-0.5em}
    \caption{\textbf{Qualitative comparison of generated videos.} Comparison across Aesthetic quality, Motion dynamics, and Prompt Following. Our VideoAWR achieves superior visual fidelity, temporal consistency, and prompt alignment compared to Flash-GRPO and Vanilla baselines.}
    \label{fig:showcase}
    \vspace{-1.5em}
\end{figure*}

\subsection{Decoupled Timestep Rectification}
\label{sec:dtr}
The behavior-anchored regression in Eq.~\eqref{eq:loss} is evaluated at a
single point $(x_t,c,t)$ and, unlike policy-gradient methods, never forms a
trajectory likelihood. This frees the training timestep from the rollout
schedule: it need not follow the sampling discretization, nor be shared across
a prompt group. Decoupled Timestep Rectification (DTR) exploits this freedom
by drawing $t_i\overset{\mathrm{iid}}{\sim}\mathcal{U}[0,1]$ for each rollout
$i=1,\dots,G$ and supervising each sample at its own timestep:
\begin{equation}
\label{eq:dtr_loss}
\widehat{\mathcal{L}}_{\mathrm{DTR}}
= \frac{1}{G} \sum_{i=1}^G
\big\| v_{\theta}(x_{t_i}^{(i)},c,t_i) - y_{t_i}^{(i)} \big\|_2^2.
\end{equation}
Assigning distinct timesteps within a group is legitimate precisely because no
term in the objective couples the temporal variable to the terminal outcome. With $t \perp (x_1, R)$, the estimator remains unbiased: $\mathbb{E}[\widehat{\mathcal{L}}_{\mathrm{DTR}}] = \mathcal{L}_{\mathrm{VideoAWR}}$ (Appendix~\ref{app:proof_dtr}). Compared to fixed-timestep sampling (eg., Flash-GRPO~\cite{he2026flash}), DTR improves temporal coverage within each batch. While single-step SDE targets typically introduce severe gradient variance, our behavior-anchored regression formulation circumvents this, yielding faster convergence and higher final performance (Figure~\ref{fig:ablation}(b)).

\paragraph{Discarding the importance correction.}
Recovering the unguided FlowAWR objective exactly would reweight each sample by
$\rho_{\mathrm{joint}}=\pold(x_1,x_t\mid c)/\qs(x_1,x_t\mid c)$. Both flows
share the forward kernel $p(x_t\mid x_1)$, so it cancels and the correction
collapses to $\rho(c,x_1)=\pold_1(x_1\mid c)/\qs(x_1\mid c)$, independent of
$t$ (Appendix~\ref{app:proof_is}). We set
$\rho\equiv1$: such a trajectory-global prefactor is exactly what DTR's
point-wise objective cannot accommodate, and any prompt-level constant is absorbed by the group normalization in
Eq.~\eqref{eq:softmax_advantage}.

\section{Experiments}
\label{sec:experiments}
\subsection{Experimental Setup}
\noindent\textbf{Datasets and Models.}
Following DanceGRPO~\cite{xue2025dancegrpo}, we use their prompt set for
training and the same split of 300 prompts in Flash-GRPO~\cite{he2026flash} for evaluation. We
build on the Wan2.1 family~\cite{wan2025}, validating VideoAWR on both the
1.3B and the 14B variants.

\noindent\textbf{Implementation Details.}
Guided rollouts use a CFG scale of $4.5$, $20$ sampling steps and group size
$G=8$ for both backbones, with an EMA
reference policy. Full training and evaluation details are in Appendix~\ref{appendix:B}.

\noindent\textbf{Baselines.}
We compare against Flow-GRPO~\cite{liu2026flow}, Flow-GRPO-Fast,
Flash-GRPO~\cite{he2026flash}, and DiffusionNFT~\cite{zheng2025diffusionnft}.
Flow-GRPO uses its official video-RL configuration, which trains only on the
first half of the denoising timesteps. As Flow-GRPO-Fast's few-step training
is conceptually aligned with MixGRPO~\cite{li2025mixgrpo}, we evaluate it with
single-step updates (Flow-GRPO-Fast1); Flash-GRPO likewise supervises one
timestep per update but shares it across the prompt group.

\noindent\textbf{Rewards and Evaluation.}
We optimize two complementary rewards. For \emph{visual quality} we use
HPSv3~\cite{ma2025hpsv3}, scoring all sampled frames and averaging the
top-$30\%$ following LongCat~\cite{team2025longcat}, which mitigates spurious
low scores caused by content inconsistency across temporal transitions. For
\emph{motion quality} we use the motion score of
VideoAlign~\cite{liu2026improving}. We additionally report
VBench~\cite{huang2024vbench} for overall quality. 

\begin{figure*}[!t]
    \centering
    \includegraphics[width=1\textwidth]{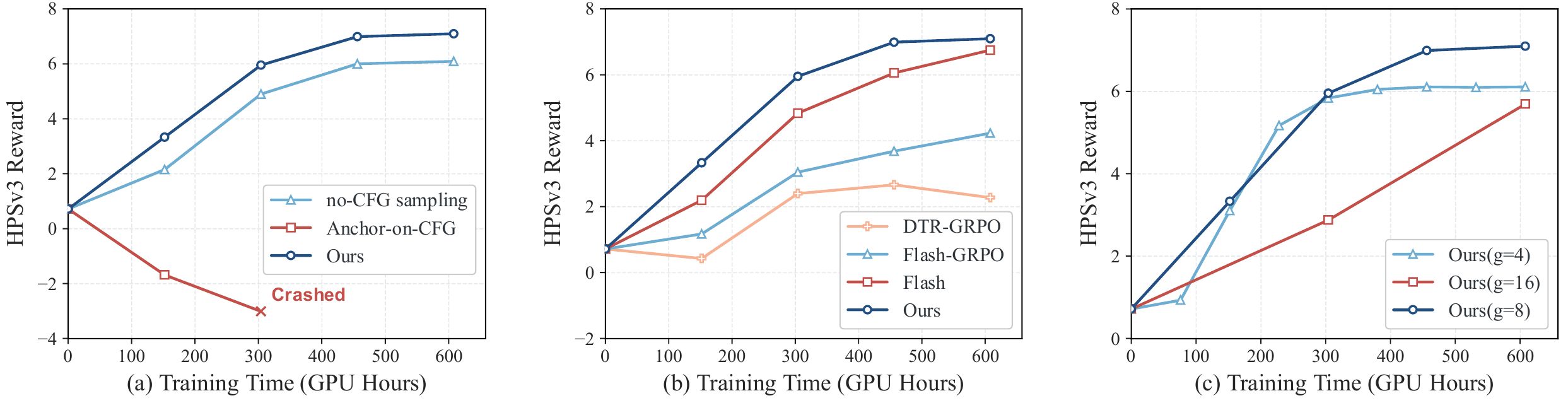}
    \vspace{-1em}
    \caption{\textbf{Ablations on Wan2.1-1.3B with the HPSv3 reward,} plotted
    against training cost in GPU hours. \textbf{(a) Rollout and anchor.} Ours
    is highest throughout; \emph{no-CFG sampling} falls behind during cold
    start and does not close the gap, while \emph{anchor-on-CFG} decreases
    monotonically and diverges before $300$ GPU hours, as predicted by
    Proposition~\ref{prop:geometric_blowup}. \textbf{(b) Timestep scheme.}
    Independent timesteps (DTR) versus a group-shared timestep (Flash), under
    our objective and under GRPO: ours exceeds GRPO at either scheme, and DTR
    helps within our objective (Ours $>$ Ours-Flash) but hurts under GRPO
    (DTR-GRPO $<$ Flash-GRPO). \textbf{(c) Group size.} At a matched GPU budget
    $G{=}8$ is the best trade-off.}
    \label{fig:ablation}
    \vspace{-1em}
\end{figure*}

\subsection{Main Results}
\label{sec:main_results}
\noindent\textbf{Reward Optimization.}
Figure~\ref{fig:main_curves} reports HPSv3 (top) and Motion Quality (bottom)
against training cost. On HPSv3 training reward (a), VideoAWR reaches $8.1$ at
$600$ GPU hours against $7.0$ for DiffusionNFT, while Flash-GRPO and
Flow-GRPO-Fast1 never exceed $4.4$. Under CFG-free inference (b), Flash-GRPO
improves fastest at first but saturates near $5.5$; VideoAWR overtakes it at
roughly $350$ GPU hours and ends at $6.75$, so the gains explored under guided
rollouts are retained by the unguided policy. Enabling CFG (c) lifts it to
$8.1$ while the policy-gradient baselines stay essentially unchanged, as
predicted in Sec.~\ref{sec:analysis}. Motion Quality (e,\,f) reproduces this
ordering, though DiffusionNFT closes most of the training-reward gap (d) by
$300$ GPU hours.

\noindent\textbf{General Video Quality.}
Table~\ref{vbench} reports VBench. VideoAWR is best on Aesthetic Quality
($69.18$), Subject Consistency ($98.48$) and Object Class ($92.53$), and second
on Imaging Quality ($71.01$ vs.\ $71.50$), at the same $600$ GPU-hour budget as
the other single-step methods. Subject consistency stays $\geq 94$ throughout,
so reward optimization preserves the structural integrity of the backbone in terms of subject consistency.

\noindent\textbf{Scaling to Wan2.1-14B.}
Figure~\ref{fig:pipeline}(a,\,b) applies the same recipe---CFG scale, group
size, clip ratio and EMA rate unchanged---to the 14B backbone under HPSv3.
Training and held-out reward both increase monotonically, reaching $9.7$ versus
$8.2$ for 1.3B on the evaluation prompts. VBench and motion quality are on 1.3B
only, where all baselines could be reproduced under a common budget.

\noindent\textbf{Qualitative Comparison.}
Relative to the vanilla Wan2.1 backbone and Flash-GRPO,
Figure~\ref{fig:showcase} shows sharper frames, more coherent motion and closer
prompt alignment.

\subsection{Ablation Study}
\label{sec:ablation}
We ablate the three design choices of VideoAWR on Wan2.1-1.3B with the HPSv3
reward.

\noindent\textbf{Guided Rollout.}
We compare our design against two alternatives with the identical objective
but different sampling and anchoring: (i) \emph{no-CFG sampling}, rolling out
without guidance; and (ii) \emph{anchor-on-CFG}, rolling out with guidance but
anchoring the target on the CFG field
(Proposition~\ref{prop:geometric_blowup}). In Figure~\ref{fig:ablation}(a),
no-CFG sampling falls behind during cold start and does not close the gap
after recovering ($6.1$ vs.\ $7.1$ at $600$ GPU hours), while anchor-on-CFG
decreases from the first checkpoint and diverges before $300$ GPU hours, as
predicted by Proposition~\ref{prop:geometric_blowup}.

\noindent\textbf{Decoupled Timestep.}
We compare independent timesteps (DTR) against a group-shared timestep
(Flash), each under our behavior-anchored objective and under a GRPO objective.
In Figure~\ref{fig:ablation}(b) both of our variants exceed both GRPO variants,
but the effect of DTR itself is not uniform: it helps under our objective
(Ours $7.1$ vs.\ Ours-Flash $6.8$) and hurts under GRPO (DTR-GRPO $2.3$ vs.\
Flash-GRPO $4.2$). This follows from Sec.~\ref{sec:dtr}: a policy-gradient
objective weights each update by an importance ratio evaluated along the sampled
trajectory, so its timestep must stay tied.

\noindent\textbf{Group Size.}
We vary $G\in\{4,8,16\}$ at a matched compute budget rather than a matched
number of updates, since $G$ scales rollout cost linearly. In
Figure~\ref{fig:ablation}(c) the trade-off is non-monotone: $G=4$ tracks $G=8$
early but saturates near $6.1$, estimating $\phi_s$ in
Eq.~\eqref{eq:softmax_advantage} too coarsely, while $G=16$ is more accurate
per update but completes too few updates within $600$ GPU hours ($5.7$). We
adopt $G=8$ as default.

\section{Conclusion}
VideoAWR formulates ``guided rollout, unguided training'' as behavior-anchored advantage-weighted rectification. By restricting guidance strictly to the sampling phase, the optimized policy isolates the reward-correlated velocity increment and filters out the guidance offset, thereby preventing the geometric divergence associated with CFG-anchored training. Consequently, the fine-tuned model supports both inference regimes: it enables efficient CFG-free single-branch generation while preserving the original semantics of inference-time CFG. Integrated with Decoupled Timestep Rectification, VideoAWR provides a stable online optimization framework that maintains better inference flexibility for both CFG-free and CFG-enabled inference.

\paragraph{Limitations and future work.}
Proposition~\ref{prop:population_optimum} mathematically permits arbitrary behavior distributions, establishing that the optimization anchor is invariant to the sampling policy. Currently, VideoAWR solely utilizes CFG-guided rollouts from the EMA backbone. Consequently, optimization is bounded by the base model's inherent generation capabilities; prompts that the backbone fundamentally fails to render cannot yield informative reward gradients, regardless of the guidance scale. Because anchor invariance supports any behavior distribution, integrating off-policy or out-of-distribution terminals requires no modification to the underlying objective. Investigating this off-policy data integration constitutes a primary direction for future work.

\bibliography{aaai2027}

\clearpage
\appendix

\section{Theoretical Derivations and Proofs}
\label{app:proofs}
\subsection{Proof of Proposition~\ref{prop:geometric_blowup} (Geometric Explosion)}
\label{app:proof_prop1}
We analyze the evolution of the conditional--unconditional velocity gap
$\Delta_k(x_t)=\vold_k(x_t,c)-\vold_k(x_t,\varnothing)$ under the
CFG-anchored variant, in which the rectification target of round $k$ is
built upon the guided field
\begin{equation*}
v^{\mathrm{cfg}}_{k}(x_t,c) =\vold_k(x_t,\varnothing)+s\,\Delta_k(x_t).
\end{equation*}
The argument operates in the linear-field regime, where the network
parameterization is expressive enough that field-space targets translate
into field-space updates to first order. This proof relies on two theoretical 
assumptions stated in the proposition: (a) within each round, the inner regression is
run to convergence, ensuring the trained conditional branch reaches its anchor;
(b) the unconditional branch receives no supervision and is treated as
invariant across rounds.

\paragraph{Step 1: within-round absorption (exact under (a)).}
By assumption (a), the trained conditional field at the end of round $k$
satisfies
\begin{equation*}
\begin{aligned}
v_{\theta_{k+1}}(x_t,c)
&=v^{\mathrm{cfg}}_{k}(x_t,c)\\
&=\vold_k(x_t,\varnothing)
  +s\,\Delta_k(x_t),
\end{aligned}
\end{equation*}
up to the advantage-weighted residual term, which is bounded by the
advantage spread and independent of $s$. It contributes only an $O(\Advs)$
perturbation that does not alter the leading-order recursion derived below.

\paragraph{Step 2: EMA merge in field space.}
This step requires a first-order linearization. The EMA update
$\told_{k+1}\leftarrow\mu_k\told_k+(1-\mu_k)\theta_{k+1}$ is strictly linear in
parameters. Linearizing the parameter-to-field map $\theta\mapsto v_\theta$
around $\told_k$ therefore transfers the same weights $\mu_k, 1-\mu_k$ to
the field space, yielding an error of order
$O(\|\theta_{k+1}-\told_k\|^2)$:
\begin{equation*}
\begin{aligned}
\vold_{k+1}(x_t,\cdot)
&\approx\mu_k\,\vold_{k}(x_t,\cdot)\\
&\quad+(1-\mu_k)\,
v_{\theta_{k+1}}(x_t,\cdot).
\end{aligned}
\end{equation*}
Applying this relationship to both branches and invoking assumption (b) for the
unconditional field, $\vold_{k+1}(x_t,\varnothing)=\vold_k(x_t,\varnothing)$,
the updated gap evaluates to
\begin{align*}
\Delta_{k+1}
&\approx\mu_k\,\Delta_k
  +(1-\mu_k)s\,\Delta_k\\
&=\big[\mu_k+(1-\mu_k)s\big]\,\Delta_k.
\end{align*}

\paragraph{Step 3: geometric compounding.}
The step-wise factor $[\mu_k+(1-\mu_k)s]$ is determined entirely by the 
pre-defined EMA schedule and the guidance scale. Because it is independent of the 
model state $\told_k$, the one-step recursive relation can be unrolled to yield
\begin{equation*}
\Delta_k \approx \left( \prod_{i=0}^{k-1} \big[\mu_i + (1-\mu_i)s\big] \right) \Delta_0,
\end{equation*}
which remains valid as long as the Step-2 linearization holds. 
For any guidance scale $s>1$ and valid EMA rate $\mu_i\in[0,1)$, the amplification 
factor strictly satisfies $\mu_i+(1-\mu_i)s>1$. Consequently, the gap exhibits 
geometric explosion. 

Because the effective guidance applied during rollout in round $k$ scales 
with $s\,\Delta_k$, the sampler operates at an exponentially increasing 
effective guidance strength. This mathematically drives the intermediate 
variables $x_t$ outside the pre-trained data manifold, which corresponds to 
the empirical training collapse. 

In contrast, anchoring the target on the unguided conditional field eliminates 
the $s$-fold absorption in Step 1. The conditional branch is regressed toward 
$\vold_k(x_t,c)$ alongside the advantage-bounded residual, reducing the recursion 
multiplier to $\mu_k+(1-\mu_k)\cdot 1=1$ at leading order. Thus, the gap evolves 
exclusively through the robust, reward-driven $O(\Advs)$ increments. $\hfill\square$

\subsection{Proof of Proposition~\ref{prop:population_optimum} (Anchor-Invariant Improvement)}
\label{app:proof_prop2}
Throughout, fix the spatio-temporal coordinate $(x_t,c,t)$; all expectations
are taken with respect to the guided posterior $\qs(x_1\mid x_t,c)$ induced
by the behavior flow, and we abbreviate
$\Exp_{\qs}[\cdot]=\Exp_{\qs(x_1\mid x_t,c)}[\cdot]$. Recall the definitions
\begin{equation*}
\begin{aligned}
\phi_s
&=\Exp_{\qs}\!\big[e^{\Rew/\gamma}\big],\\
\Advs
&=\frac{e^{\Rew/\gamma}}{\phi_s}-1,\\
\vs
&=\Exp_{\qs}[\uvec],\\
\vsstar
&=\vs+\Exp_{\qs}\big[\Advs(\uvec-\vs)\big].
\end{aligned}
\end{equation*}

\paragraph{Step 1: the population minimizer is a conditional expectation.}
The behavior-anchored population objective stated in the main paper regresses
$\vt(x_t,c,t)$, a deterministic function of the network input, onto the
stochastic target
$y=\vold+\Advs(\uvec-\vold)$, whose residual randomness given the input is
carried entirely by $x_1\sim\qs(\cdot\mid x_t,c)$. By the standard
$L^2$-projection property of the conditional expectation, the pointwise
minimizer is
\begin{equation}
\begin{aligned}
v^{\dagger}
&=\Exp_{\qs}\big[
  \vold+\Advs(\uvec-\vold)
  \big]\\
&=\vold+
  \Exp_{\qs}\big[\Advs(\uvec-\vold)\big],
\end{aligned}
\label{eq:app_vdagger}
\end{equation}
where the second equality uses that $\vold$ is constant with respect to the
integration variable $x_1$ at fixed $(x_t,c,t)$.

\paragraph{Step 2: zero mean of the centered advantage.}
Directly from the definitions,
\begin{equation}
\begin{aligned}
\Exp_{\qs}[\Advs]
&=\frac{
  \Exp_{\qs}\!\big[e^{\Rew/\gamma}\big]
  }{\phi_s}-1\\
&=\frac{\phi_s}{\phi_s}-1=0.
\end{aligned}
\label{eq:app_zero_mean}
\end{equation}
Note that this identity holds precisely because $\phi_s$ is the value
function of the \emph{guided} posterior---the same measure under which the
expectation is taken. Consequently, for \emph{any} vector $w(x_t,c,t)$
independent of $x_1$,
\begin{equation}
\begin{aligned}
\Exp_{\qs}\big[\Advs\,w\big]
&=w\cdot\Exp_{\qs}[\Advs]\\
&=0.
\end{aligned}
\label{eq:app_anchor_invisible}
\end{equation}

\paragraph{Step 3: anchor exchange.}
Applying Eq.~\eqref{eq:app_anchor_invisible} with $w=\vold$ and then with
$w=\vs$,
\begin{equation*}
\begin{aligned}
\Exp_{\qs}\big[\Advs(\uvec-\vold)\big]
&=\Exp_{\qs}\big[\Advs\,\uvec\big]\\
&=\Exp_{\qs}\big[\Advs(\uvec-\vs)\big],
\end{aligned}
\end{equation*}
i.e., the anchor subtracted inside the residual is immaterial in
expectation. Equivalently, since $\Advs$ is centered, the improvement term
is a covariance,
$\Exp_{\qs}[\Advs\,\uvec]=\mathrm{Cov}_{\qs}(\Advs,\uvec\mid x_t,c)$, which
is invariant to constant shifts of $\uvec$; the guidance offset
$(\vs-\vold)$ is exactly such a shift at fixed $x_t$.

\paragraph{Step 4: identification with the guided-flow optimum.}
By the definition of $\vsstar$,
\begin{equation*}
\begin{aligned}
\Exp_{\qs}\big[\Advs(\uvec-\vs)\big]
&=\vsstar-\vs.
\end{aligned}
\end{equation*}
Substituting into Eq.~\eqref{eq:app_vdagger} yields
\begin{equation*}
\begin{aligned}
v^{\dagger}
&=\vold+\big(\vsstar-\vs\big),\\
v^{\dagger}-\vsstar
&=\vold-\vs,
\end{aligned}
\end{equation*}
completing the proof. $\hfill\square$

\subsection{Unbiasedness of Decoupled Timestep Rectification}
\label{app:proof_dtr}
The behavior-anchored population loss stated in the main paper is an expectation over the
product measure of $t\sim\mathcal{U}[0,1]$, $x_0\sim p_0$, and
$x_1\sim\qs(\cdot\mid c)$; denote the integrand by
$\ell_\theta(t,x_0,x_1)=\|\vt(x_t,c,t)-y_t\|_2^2$, and abbreviate
\begin{equation*}
\ell_\theta^{(i)}
:=\ell_\theta(t_i,x_0^{(i)},x_1^{(i)}).
\end{equation*}
In DTR, each rollout
$i$ draws $t_i\sim\mathcal{U}[0,1]$ independently of $(x_0^{(i)},
x_1^{(i)},\Rew^{(i)})$, so the tuple $(t_i,x_0^{(i)},x_1^{(i)})$ is
distributed according to the same product measure. Hence, for each $i$,
\begin{equation*}
\begin{aligned}
\Exp\big[\ell_\theta^{(i)}\big]
&=\mathcal{L}_{\mathrm{VideoAWR}}(\theta),
\end{aligned}
\end{equation*}
and by linearity of expectation over the group average,
\begin{equation*}
\begin{aligned}
\Exp\big[\widehat{\mathcal{L}}_{\mathrm{DTR}}\big]
&=\frac{1}{G}\sum_{i=1}^{G}
\Exp\big[\ell_\theta^{(i)}\big]\\
&=\mathcal{L}_{\mathrm{VideoAWR}}(\theta).
\end{aligned}
\end{equation*}
The independence $t\perp(x_1,\Rew)$ provides the theoretical justification for distinct
timesteps to distinct rollouts: no term in the objective couples the
temporal variable to the terminal reward, unlike trajectory-likelihood
methods in which the training timestep must follow the discretization of
the sampled trajectory. $\hfill\square$

\subsection{Cancellation of the Forward Kernel in the Importance Weight}
\label{app:proof_is}
Both the reference flow and the guided behavior flow share the identical
forward noising process: given a terminal sample $x_1$, the intermediate
state is constructed as $x_t=(1-t)x_0+t\,x_1$ with $x_0\sim p_0$,
independently of which policy generated $x_1$. Their joint densities over
$(x_1,x_t)$ therefore factorize with a common conditional kernel:
\begin{equation*}
\begin{aligned}
\pold(x_1,x_t\mid c)
&=\pold_1(x_1\mid c)\,
  p(x_t\mid x_1),\\
\qs(x_1,x_t\mid c)
&=\qs(x_1\mid c)\,
  p(x_t\mid x_1).
\end{aligned}
\end{equation*}
Define the terminal density ratio
\begin{equation*}
\rho(c,x_1)
:=\frac{\pold_1(x_1\mid c)}
        {\qs(x_1\mid c)}.
\end{equation*}
Taking the joint ratio, the common kernel cancels:
\begin{equation*}
\begin{aligned}
\rho_{\mathrm{joint}}(x_1,x_t)
&=\rho(c,x_1)
  \frac{p(x_t\mid x_1)}
       {p(x_t\mid x_1)}\\
&=\rho(c,x_1),
\end{aligned}
\end{equation*}
which depends only on the prompt--terminal pair and is constant along the
entire noising trajectory, in particular independent of the training
timestep $t$. $\hfill\square$

\section{Implementation Details}
\label{appendix:B}

\paragraph{Backbones and Data.}
We fine-tune the official Diffusers checkpoints of Wan2.1-T2V-1.3B and Wan2.1-T2V-14B~\citep{wan2025}. All videos are generated at $480\times832$ resolution with $81$ frames at $16$ FPS. The text encoder processes a maximum of $512$ tokens. Training is conducted in \texttt{bfloat16} mixed precision.

\paragraph{Parameter-Efficient Fine-Tuning.}
We employ rank-$32$ LoRA adapters ($\alpha=64$, Gaussian initialization) on the query, key, value, and output projections of both the standard and newly introduced attention blocks, while keeping the base network frozen.

\begin{table*}[!t]
    \centering
    \caption{\textbf{CFG-free and CFG-enabled evaluation of the official
    DiffusionNFT checkpoint.} All metrics are higher-is-better. Best values
    between the two inference modes are in \textbf{bold}.}
    \label{tab:diffusionnft_cfg_compatibility}
    \small
    \setlength{\tabcolsep}{3.5pt}
    \begin{tabular}{@{}l ccc ccccc@{}}
      \toprule
      \textbf{Method}
      & \multicolumn{3}{c}{\textbf{Task Metric}}
      & \multicolumn{5}{c}{\textbf{DrawBench Metric}} \\
      \cmidrule(lr){2-4} \cmidrule(lr){5-9}
      & \textbf{GenEval} & \textbf{OCR} & \textbf{PickScore}
      & \textbf{CLIPScore} & \textbf{HPSv2.1} & \textbf{Aesthetic}
      & \textbf{ImgRwd} & \textbf{PickScore} \\
      \midrule
      DiffusionNFT (CFG-free)
      & \textbf{0.93} & \textbf{0.90} & \textbf{23.36}
      & \textbf{0.293} & \textbf{0.330} & \textbf{5.99}
      & \textbf{1.48} & \textbf{23.79} \\
      DiffusionNFT (with CFG)
      & 0.84 & 0.69 & 20.99
      & 0.291 & 0.273 & 5.16
      & 1.03 & 22.03 \\
      \bottomrule
    \end{tabular}
\end{table*}

\begin{figure*}[!t]
    \centering
    \includegraphics[width=0.98\textwidth]{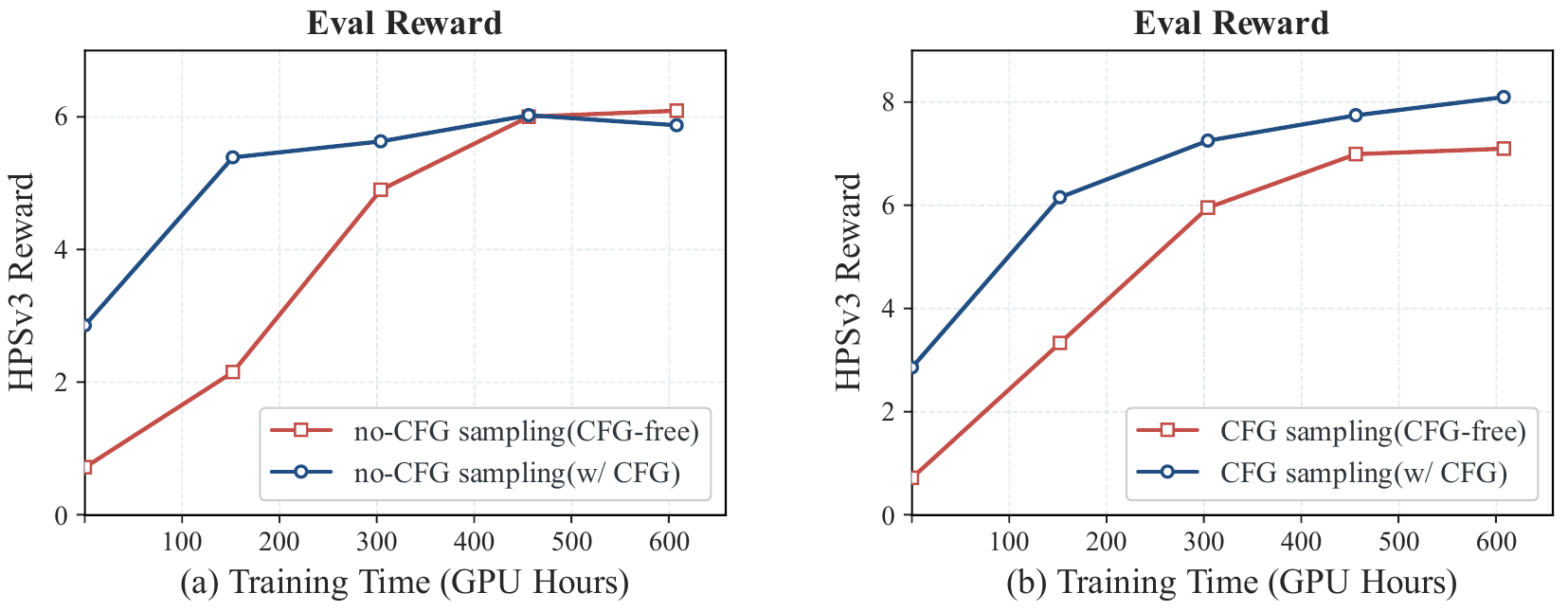}
    \vspace{-1em}
    \caption{\textbf{CFG-free and CFG-enabled evaluation under different
    behavior samplers.} We train Wan2.1-T2V-1.3B with HPSv3 while anchoring
    the rectification target on the unguided conditional field
    $v^{\mathrm{old}}$ in both panels. \textbf{(a)} No-CFG behavior sampling.
    \textbf{(b)} CFG-guided behavior sampling. Each trained checkpoint is
    evaluated both CFG-free (red) and with CFG (blue). Guided behavior raises
    both curves and preserves a consistent inference-time CFG gain.}
    \label{fig:appendix_cfg_compatibility}
    \vspace{-1em}
\end{figure*}

\paragraph{Rollout and Sampling.}
Guided rollouts are generated using the frozen behavior policy adapter. We use the native flow-matching scheduler with $20$ integration steps and a CFG scale of $s=4.5$. The primary configuration uses a prompt group size of $G=8$. A distributed repeat sampler ensures all $G$ rollouts for a single prompt are constructed synchronously. For Decoupled Timestep Rectification (DTR), we align with Flash-GRPO~\citep{he2026flash} and restrict single-step supervision to the first $51\%$ of the rollout schedule (i.e., candidate indices $\{0, \ldots, 9\}$ for a 20-step schedule). For each sample, the training timestep index is drawn uniformly and independently.

\paragraph{Reward Configuration.}
Video quality is evaluated using the HPSv3 reward~\citep{ma2025hpsv3}. We score individual frames and average the top $30\%$ to mitigate spurious variance at temporal transitions~\citep{team2025longcat}. For each prompt group of $G$ rollouts sampled from the guided behavior distribution, we compute the group-centered advantage
\begin{equation*}
    \Advs^{(i)} = G\frac{\exp(R_i/\gamma)}{\sum_{j=1}^{G}\exp(R_j/\gamma)} - 1,
\end{equation*}
where $R_i$ is the reward of the $i$-th rollout and $\gamma>0$ is the temperature. This is the group estimate of $\Advs$, computed as $G$ times the group-wise Softmax weight minus one. The advantages have zero mean within each group and can be positive, negative, or zero. Zero-advantage filtering is disabled.

\paragraph{Loss Formulation.}
The training utilizes the clean-latent prediction formulation. Defining the model's clean latent estimate as $\widehat z_i^{\,\theta} = x_{\tau_i}^{(i)} - \tau_i v_\theta$ (where $\tau_i$ denotes the forward time parameterization), the unguided behavior reference as $\widehat z_i^{\,\mathrm{old}}$, and the base-model reference as $v_{\mathrm{ref}}$, the target latent is constructed as $z_i^{\mathrm{tgt}} = \widehat z_i^{\,\mathrm{old}} + \Advs^{(i)} (z_i - \widehat z_i^{\,\mathrm{old}})$. The final objective combines the rectified policy loss with a KL penalty:
\begin{equation*}
\begin{split}
    \mathcal{L}_{\mathrm{rect}} &= \mathbb{E}_i \left[ \frac{\left( \widehat z_i^{\,\theta} - z_i^{\mathrm{tgt}} \right)^2}{\operatorname{sg}\left(\max\{ \mathbb{E}[|\widehat z_i^{\,\theta} - z_i|], 10^{-5} \}\right)} \right], \\
    \mathcal{L}_{\mathrm{KL}} &= \mathbb{E} \left[ \| v_\theta - v_{\mathrm{ref}} \|_2^2 \right], \\
    \mathcal{L} &= \mathcal{L}_{\mathrm{rect}} + 0.0001 \mathcal{L}_{\mathrm{KL}},
\end{split}
\end{equation*}
where $\operatorname{sg}(\cdot)$ denotes the stop-gradient operator acting as an adaptive normalizer.

\paragraph{Optimization and Hardware.}
Training is executed on a cluster of $48$ NVIDIA H200 GPUs. The 1.3B model utilizes DDP, while the 14B model employs FSDP (\texttt{FULL\_SHARD}) with activation checkpointing. We optimize the LoRA parameters using AdamW with a learning rate of $10^{-4}$, $\beta = (0.9, 0.999)$, and weight decay of $10^{-4}$. The maximum gradient norm is clipped at $1.0$, and the numerical stability term is $\epsilon=10^{-8}$.

\paragraph{Policy Synchronization and Evaluation.}
The behavior adapter ($\theta^{\mathrm{old}}$) is synchronized from the trainable policy via a scheduled EMA update: exact copies for the first $75$ steps, a linear transition up to step $208$, and an EMA decay of $\lambda = 0.999$ thereafter. Separately, an evaluation EMA with a $0.9$ decay is maintained over the trainable parameters for checkpointing and inference.

\section{Validation of CFG-Free and CFG-Enabled Inference}
\label{appendix:cfg_compatibility}

\paragraph{Why a sampling-distribution anchor can impair CFG.}
Anchoring a regression target directly on the guided sampling field provides
a straightforward route to CFG-free inference: the conditional branch
internalizes the amplified behavior field and can therefore generate strong
samples without evaluating an unconditional branch. However, this changes the
semantics of the conditional--unconditional gap. Applying CFG again may
extrapolate from a conditional field that has already absorbed guidance,
effectively amplifying the same offset twice.

We first examine this effect using the official DiffusionNFT
checkpoint~\citep{zheng2025diffusionnft}. As shown in
Table~\ref{tab:diffusionnft_cfg_compatibility}, the released model is
strong under CFG-free inference, but enabling CFG decreases every reported
Task and DrawBench metric. The largest drops occur on OCR
($0.90\rightarrow0.69$), HPSv2.1 ($0.330\rightarrow0.273$), ImageReward
($1.48\rightarrow1.03$), and the two PickScore evaluations
($23.36\rightarrow20.99$ and $23.79\rightarrow22.03$). This consistent
degradation is compatible with guidance having been absorbed into the learned
conditional field, although the comparison alone does not identify the
mechanism causally.

This observation motivates the dual-mode requirement targeted by VideoAWR:
guided behavior should improve exploration without being baked into the
conditional anchor. The resulting policy should retain the learned
reward-driven improvement under CFG-free inference while preserving a
meaningful conditional--unconditional gap for optional inference-time CFG.

\paragraph{Video flow alignment.}
We next evaluate the same dual-mode requirement on Wan2.1-T2V-1.3B optimized
with HPSv3. Figure~\ref{fig:appendix_cfg_compatibility} isolates the behavior
sampling distribution while holding the optimization anchor fixed to the
unguided conditional field $v^{\mathrm{old}}$ in both settings. Panel (a)
uses no-CFG behavior rollouts. Although inference-time CFG provides a large
initial advantage, this benefit gradually vanishes during training. After
convergence, applying CFG no longer improves the policy and instead yields a
slightly lower evaluation reward than CFG-free inference.

Panel (b) uses guided behavior rollouts while retaining the same unguided
conditional anchor. Both inference modes improve faster and reach higher
rewards than their counterparts in panel (a). Moreover, the CFG-enabled curve
remains above the CFG-free curve as training converges. Thus, guided behavior
improves the single-branch policy without requiring the guided field to enter
the regression anchor, while preserving the benefit of optional
inference-time CFG after training.

\paragraph{Connection to Proposition~\ref{prop:population_optimum}.}
Proposition~\ref{prop:population_optimum} characterizes the population minimizer as
\begin{equation*}
v^{\dagger}
=v^{\mathrm{old}}+(v_s^*-v_s),
\end{equation*}
namely, the unguided conditional anchor plus the reward-correlated improvement
explored under the guided behavior distribution. The behavior guidance offset
$v_s-v^{\mathrm{old}}$ is absent from the learned backbone. The results in
Figure~\ref{fig:appendix_cfg_compatibility}(b) are consistent with both
consequences of this characterization: the guided rollouts yield a stronger
CFG-free policy, while the retained gain from inference-time CFG indicates
that the conditional--unconditional gap has not been absorbed into the
conditional anchor. The figure therefore provides empirical support for the
structural prediction of Proposition~\ref{prop:population_optimum}; the exact identity itself follows
from the population argument in Appendix~\ref{app:proof_prop2}.

\begin{figure*}[!t]
    \centering
    \includegraphics[width=0.99\textwidth]{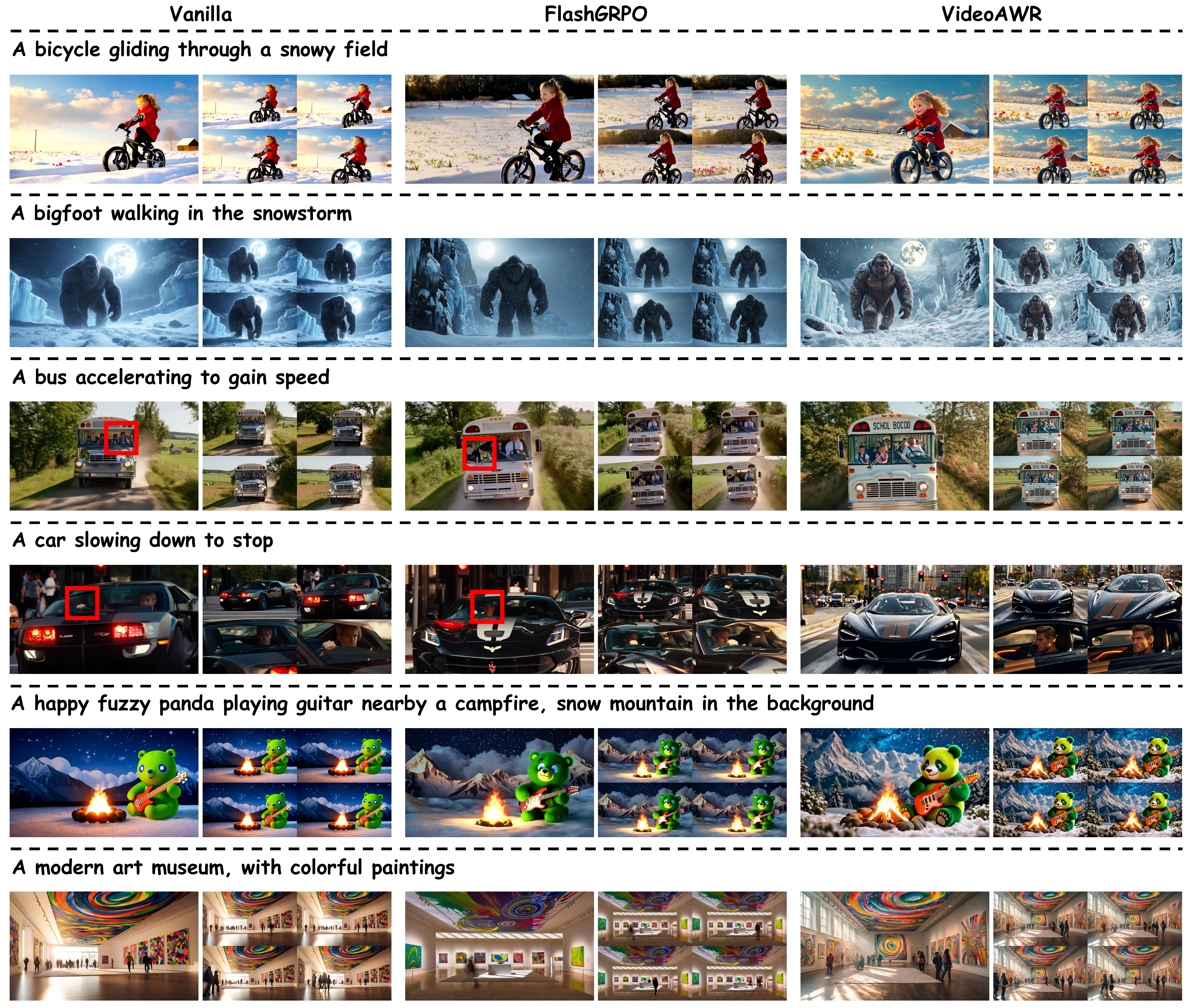}
    \caption{Qualitative comparison among VideoAWR, Flash-GRPO, and the Vanilla baseline fine-tuned with the HPSv3 reward on VBench prompts (Wan14B).}
    \label{fig:appendix_case_Wan14B}
    \vspace{-1.2em}
\end{figure*}

\section{More Qualitative Evaluation}
\label{appendix:D}

We qualitatively compare vanilla Wan2.1 with its Flash-GRPO and VideoAWR aligned variants at the 14B model scales. As shown in Figure~\ref{fig:appendix_case_Wan14B}, VideoAWR consistently generates videos with higher visual fidelity, richer scene details, and smoother motion dynamics.

As shown in Figure~\ref{fig:appendix_case_Wan14B}, VideoAWR shows consistent improvements across diverse scenes. In the animated scenes, the little girl riding a bicycle and Bigfoot walking through the snow exhibit more plausible motion dynamics and higher aesthetic quality. The scenes depicting a middle-aged driver speeding along a rural road in a school bus and a blue-eyed man decelerating a black sports car demonstrate greater semantic consistency and more coherent motion. The scene featuring a green-furred panda playing a guitar beside a campfire against a snow-covered mountain backdrop presents richer visual details and an enhanced depth-of-field effect. Finally, the scene depicting a modern art museum displaying abstract paintings exhibits higher visual fidelity and a better organized spatial layout.

These results confirm that VideoAWR effectively improves visual aesthetics, temporal coherence, and prompt adherence across different model scales and content types.

\noindent \textbf{Prompts in Figure~\ref{fig:showcase}. } The whole prompts in Figure~\ref{fig:showcase} are as follows:
\begin{enumerate}
\item CG animation concept art, a vintage steam locomotive slowly traveling down a narrow railway track. The train is adorned with intricate wooden carvings and painted with vibrant floral patterns. Smoke billows out from the chimney, creating a warm and nostalgic atmosphere. The engine driver, a rugged man with a mustache, stands at the front, looking determinedly forward. He wears a leather hat and a sturdy coat. The scenery along the track consists of lush greenery, small villages, and winding paths. The train moves at a leisurely pace, allowing passengers to enjoy the scenic views. The lighting is soft and golden, casting a gentle glow over everything. Low-angle view, focusing on the engine and the conductor, with medium shot of the interior and exterior of the train.
\item A sleek black motorcycle weaving skillfully through a narrow city street corner. The rider, clad in tight leather gear, leans confidently into the turn with their helmet securely fastened. The motorcycle's tires screech lightly as it navigates the tight curve, showcasing impressive handling skills. The background features bustling traffic and illuminated buildings, adding to the dynamic atmosphere. The scene captures the essence of urban racing and adrenaline-pumping moments. Low-angle shot focusing on the rider and motorcycle, medium shot of the corner, and a wide shot of the intersection.
\item Cyberpunk-inspired visual style, a towering super cool giant robot standing in the heart of futuristic Cyberpunk Beijing. The robot has sleek, metallic body with glowing neon lights and intricate circuitry patterns. It stands tall amidst the bustling cityscape, towering over skyscrapers and neon-lit streets. The background features a blend of retro and modern elements, with neon signs, flying cars, and towering skyscrapers. The robot's eyes glow intensely with a fierce determination, as it scans the city with its advanced sensors. It holds a large energy weapon in one hand, ready to defend the city from any threats. The robot's mechanical limbs move smoothly, reflecting the high-tech and futuristic atmosphere. The scene captures a moment of intense action, with people rushing by in various directions. The lighting is dim, with soft shadows highlighting the robot's imposing presence. The overall feel is gritty and exhilarating, showcasing the power and resilience of advanced technology in a dystopian future. Medium shot focusing on the robot's face and arm, capturing its intense gaze and dynamic movement.
\end{enumerate}

\noindent \textbf{Prompts in Figure~\ref{fig:appendix_case_Wan14B}. } The prompts in Figure~\ref{fig:appendix_case_Wan14B} are as follows:
\begin{enumerate}
\item A whimsical winter scene captured in high-definition, a sleek black bicycle gracefully glides through a pristine snowy field. The snow-covered ground sparkles under the soft glow of a warm sunset. The bicycle has intricate patterns and accents in silver and gold, reflecting the sunlight beautifully. A young girl, with wavy blonde hair tied into a ponytail, is riding the bike, her expression filled with joy and excitement. She wears a cozy red coat and matching mittens, and a pair of stylish black boots. The field is dotted with small, colorful wildflowers, adding a pop of vibrant hues against the white backdrop. The sky is painted with soft, fluffy clouds, hinting at a gentle breeze. The background showcases a blend of natural and artificial elements, including a few scattered wooden fences and a quaint wooden barn in the distance. The camera moves smoothly from a wide angle capturing the vast expanse of the snowy field to a closer perspective on the girl and the bicycle, emphasizing their dynamic interaction. The overall scene exudes a sense of tranquility and adventure, perfect for an animated winter tale.
\item A massive bigfoot creature with shaggy fur and powerful limbs, walking slowly and cautiously through a deep snowstorm. The bigfoot stands approximately 7 feet tall, with a broad chest and muscular build. It moves with deliberate steps, its large feet sinking slightly into the heavy snow. The creature's eyes are glowing with a fierce, determined light, scanning the surroundings for any signs of danger. Snowflakes swirl around it, creating a dense and ominous atmosphere. The bigfoot's fur is matted with frozen droplets and ice crystals, adding to its rugged appearance. In the background, towering snowdrifts and jagged icicles contrast sharply against the pale moonlight. The scene captures the bigfoot's silent progress through the treacherous winter landscape, with occasional bursts of movement as it shifts its weight or adjusts its balance. The lighting is stark and dramatic, emphasizing the bigfoot's imposing presence in the midst of the storm. Wide shot of the bigfoot in the middle of the snowy wilderness, with medium shots of its feet sinking into the snow and close-ups of its face and eyes.
\item A vintage school bus speeding down a dusty rural road, its chrome accents reflecting sunlight as it gains momentum. The bus is filled with excited children, their backpacks slung over their shoulders and faces alight with anticipation. The driver, a middle-aged man with a stern but kind expression, gazes out the front window with determination. The children lean forward, holding onto the straps, while others sit quietly, lost in their own thoughts. The bus makes sharp turns around curves, the tires screeching slightly. Trees and fields blur past the windows as the bus accelerates rapidly. The countryside stretches out before them, dotted with small houses and livestock. Aerial shot of the speeding bus from above, capturing the blur of the landscape and the determined face of the driver. Close-up shots of the children's faces, focusing on their expressions of wonder and excitement. Medium shot of the bus interior, showing the rows of seats filled with eager passengers. Low-angle shot of the bus as it speeds away, emphasizing the speed and power.
\item A sleek black sports car with a racing stripe on its hood, gradually slowing down as it approaches a red traffic light. The driver, a ruggedly handsome man with short sandy-blonde hair and intense blue eyes, grips the steering wheel tightly. His expression shows determination and focus. The car's interior is modern and luxurious, with dark wood trim and high-end leather seats. The passenger seat remains empty. As the car begins to decelerate, it emits a slight humming noise from its engine. The traffic light turns red, casting a warm amber glow over the scene. The surroundings include a busy city street with bustling pedestrians and vehicles, reflecting the urban atmosphere. The camera moves from a wide shot of the car coming to a halt, then zooms in to capture the moment the car comes to a complete stop, highlighting the car's distinctive design and the driver's focused expression. The final image is a close-up of the driver's face, captured at a low angle, with the car's taillights illuminating the scene. High dynamic range lighting, cinematic lighting. Medium shot, close-up, low-angle view.
\item A joyful fuzzy panda playing a small electric guitar near a cozy campfire. The panda has bright green fur and twinkling blue eyes, surrounded by a warm glow from the flames. In the background, majestic snow mountains with soft, fluffy clouds contrast against a clear, starry night sky. The campfire creates a soft, flickering light, casting shadows on the snow-covered ground. The panda is positioned between the fire and the mountains, its arms gracefully moving as it strums the guitar. The scene is captured with a low-angle shot, emphasizing the serene atmosphere and the lively interaction between the panda and the campfire. Snowflakes gently fall, adding a touch of winter magic. Warm ambient lighting enhances the overall mood.
\item Modern art museum interior, vibrant with colorful abstract paintings adorning the walls. Soft lighting casts a warm glow over the space, highlighting the bold brushstrokes and vibrant hues. Visitors wander among the works, captivated by the dynamic compositions. The center piece is a large mural painted on the ceiling, featuring swirling patterns and splashes of color. Surrounding it are smaller pieces arranged in a spiral pattern, each piece more striking than the last. The floor is covered in a soft, textured rug, adding depth and contrast to the surrounding colors. Natural sunlight filters through large glass windows, creating an ethereal effect. The overall atmosphere is one of excitement and wonder, inviting viewers to explore and engage with the art.
\end{enumerate}

\end{document}